\documentclass[conference]{IEEEtran}
\IEEEoverridecommandlockouts
\usepackage{cite}
\usepackage{amsmath,amssymb,amsfonts}
\usepackage{algorithmic}
\usepackage{graphicx}
\usepackage{textcomp}
\usepackage{xcolor}
\usepackage{amsmath}
\usepackage{mathtools}
\usepackage{romannum}

\usepackage{pgfplots}
\usepackage{filecontents}
\usepackage{adjustbox}

\newcommand{\norm}[1]{\left\lVert#1\right\rVert}

\def\BibTeX{{\rm B\kern-.05em{\sc i\kern-.025em b}\kern-.08em
    T\kern-.1667em\lower.7ex\hbox{E}\kern-.125emX}}
    
\begin{document}

\title{Semi-supervised Grasp Detection by Representation Learning in a Vector Quantized Latent Space
}

\author{\IEEEauthorblockN{Mridul Mahajan, Tryambak Bhattacharjee, Arya Krishnan, Priya Shukla and  G C Nandi\\}
\IEEEauthorblockA{\textit{Center of Intelligent Robotics} \\
\textit{Indian Institute of Information Technology, Allahabad, INDIA-211015}\\
priyashuklalko@gmail.com, mridulmahajan44@gmail.com and gcnandi@iiita.ac.in}}


\maketitle

\begin{abstract}

For a robot to perform complex manipulation tasks, it is necessary for it to have a good grasping ability. However, vision based robotic grasp detection is hindered by the unavailability of sufficient labelled data. Furthermore, the application of semi-supervised learning techniques to grasp detection is under-explored. In this paper, a semi-supervised learning based grasp detection approach has been presented, which models a discrete latent space using a Vector Quantized Variational AutoEncoder (VQ-VAE). To the best of our knowledge, this is the first time a Variational AutoEncoder (VAE) has been applied in the domain of robotic grasp detection. The VAE helps the model in generalizing beyond the Cornell Grasping Dataset (CGD) despite having a limited amount of labelled data by also utilizing the unlabelled data. This claim has been validated by testing the model on images, which are not available in the CGD. Along with this, we augment the Generative Grasping Convolutional Neural Network (GGCNN) architecture with the decoder structure used in the VQ-VAE model with the intuition that it should help to regress in the vector-quantized latent space. Subsequently, the model performs significantly better than the existing approaches which do not make use of unlabelled images to improve the grasp.
\end{abstract}


\section{Introduction}
The advancements in the field of automation has led to an explosive expansion in the use of intelligent machines in various applications. But even with such advancements, robots have not yet become a general-purpose utility as a whole. The reason being the ever-changing environment which calls for the need of tremendous adaptive ability and a near-perfect sense of objectivity. A process such as grasping an object, which may easily come to human beings, is a rather complex process when applied to machines. That being said, the ultimate problem boils down to an intact sense of object detection and grasping.

Earlier, tasks such as this were done using analytical approaches, which involved hard-coding the instructions involving the robot’s parameters and its world coordinates. These algorithms, called control algorithms, involved defining control over the robot’s joints \cite{kumra2017robotic} and were designed using the knowledge of human experts. These manually planned approaches achieve efficiency but are restricted by the programmer’s predictions of the robot's environment and also by dynamic environments \cite{4}. The more dynamic the actuator of the robot is intended to be, the more impossible the task becomes of physically planning it. Hence, manual teaching is efficient but exhaustive \cite{morrison2018closing}. Recently however, deep learning has remarkably advanced computer vision in fields such as classification, localisation and detection. It has also been observed that the application of computer vision to the problem of object grasping is analogous to object detection. \cite{arc1,lenz2015deep}. Hence, in most previous studies, grasp detection has been presented as a computer vision problem. Owing to an abundance of unlabelled data and an unavailability of sufficient labelled data, we want a model to utilize the unlabelled data too, hence avoiding the expense of a large labelled dataset. Also, the neural network should have the ability to generalize, so that it understands the data and the semantics behind it instead of learning only a mapping from the input space to the output space.



The major contributions of our work are listed as:
\begin{itemize}
    \item We present a semi-supervised learning based model for vision based robotic grasp detection. The model uses VQ-VAE for regressing in a vector quantized latent space. We show that by utilizing the unlabelled data, the model gives good results despite being trained on limited amount of labelled data. Along with this, we validate the generalization ability of the model as well.
    \item We augment the robotic grasp detection model's architecture with the decoder structure for better utilization of the vector-quantized latents.
\end{itemize}

Rest of the paper is organized as follows: Section \Romannum{2} analyses previous research in vision based robotic grasp detection along with highlighting the shortcomings in existing approaches. Section \Romannum{3} gives a primer on VAE, VQ-VAE and GGCNN, which are important components in our proposed approach, and Section \Romannum{4} presents the results with their analysis. Section \Romannum{5} concludes the paper along with giving hints for future research directions.

\section{Analysis of previous research}
Grasp prediction techniques can be primarily classified into two types: Analytical and Empirical. The analytical techniques involve complex models of geometry, dynamics and kinematics to determine the grasps. \cite{bicchi2000robotic} reviews such approaches in detail. However, these approaches are not always preferred due to the underlying complexity as well as the difficulty in modelling them in the real world. On the other hand, empirical techniques involve estimation models and experience-based approaches, the likes of which are discussed below.

Substantial research with the use of deep CNNs has been investigated in order to predict grasps on objects. From a broad perspective, grasp detection techniques in literature using deep learning can be divided into two main categories \cite{caldera2018review}:
\begin{enumerate}
    \item Approaches that involve designing an application-specific model. \cite{lenz2015deep, morrison2018closing, ku2017associating}
    \item Techniques which use a pre-existing model and apply it to grasp detection through transfer learning approaches. \cite{redmon2015real, watson2017real, kumra2017robotic}
\end{enumerate}
Most of them includes a two-stage pipeline \cite{lenz2015deep, wei2017robotic, wang1999lagrangian}: firstly, several grasp candidates are sampled from the image, which are then fed as inputs to a CNN network to figure out the best among the sampled candidates. This leads to a substantial execution time causing the grasps to be executed in open loop, meaning once a grasp has been determined, the robot executes it in a fixed way without taking any feedback from the environment, such as any possible changes in the location or orientation of the object after grasp determination.

\cite{lenz2015deep} uses a two-stage cascaded approach, along with the sliding window concept, to sample a number of grasp candidates. Unlike the two-staged approach, \cite{redmon2015real} and  \cite{kumra2017robotic} use adapted versions of AlexNet\cite{krizhevsky2012imagenet} and ResNet\cite{he2016deep} respectively, and are popular object detection models to make grasp predictions. Both of the previously proposed models use a single deep CNN network to regress the final grasp rectangle directly. Directly regressing a \textit{single} grasp rectangle for the entire scene might average all possible grasp rectangles that exist for an object, which itself might not be a good grasp candidate. This is because of the fact that simple averaging does not consider the spatial features of the object to be grasped.

However, \cite{morrison2018closing} possibly addresses both problems of averaging and execution time. The model, named Generative Grasp CNN or simply GGCNN, predicts grasps along with their quality for every single pixel of the image. Unlike \cite{ku2017associating}, GGCNN does not impose a constraint on the objects' shape and is fast enough to be used in real-time applications.

To the best of our knowledge, all the existing models do not consider the fact that the labelled training data for grasp detection is limited. Due to unavailability of sufficient labelled data to train a neural network for vision based grasp detection, the need to explore semi-supervised learning domain becomes evident. Also, the neural network should have the ability to generalize. Hence, we explore these ideas in detail in the next sections.


\section{Methodology}
\subsection{Preliminaries}
\subsubsection{Variational Autoencoder (VAE) \cite{2019arXiv190602691K}}
Explicit modelling of a distribution over the unlabelled training data can uncover latent representations which may help the supervised learning task of grasp detection. Variational Autoencoders (VAEs) can be used for approximate density estimation of the training data. Essentially, we first define a density function for the training data over the latent variables \textbf{z}, which is intractable to be computed for every \textbf{z}.

\begin{equation}
\mathbf{p_\theta(x)} = \int_{z} \mathbf{p_\theta(z)p_\theta(x|z)dz}
\end{equation} 

Assuming that the training data has been generated from the unobserved latent \textbf{z}, for the generation task, we sample \textbf{z} from the true prior \textbf{p(z)} and then sample \textbf{x} from the true conditional distribution over the latent \textbf{z}, ${p_\theta(x|z)}$. A Gaussian prior is generally used for its simplicity. ${p_\theta(x|z)}$ is modelled by a neural network whose optimal parameters are obtained by maximizing the likelihood of the training data. However, this data likelihood is intractable as has been discussed before. As a consequence, the posterior distribution becomes intractable as well. Hence, we define ${q_\phi(z|x)}$ as an approximation to the true posterior. This helps in deriving the lower bound for the data likelihood. 

The encoder network here is used for inference, and the decoder network is used for generation. Both of these networks are probabilistic, and therefore, output the mean and the diagonal covariance of the density function they model. The Evidence Lower Bound (ELBO) is a tractable lower bound of the likelihood of the training data whose gradient can be computed by the reparameterization trick to optimize it.
\begin{equation}
ELBO = \mathbf{E_z\big[\log p_\theta(x|z)\big]  -  D_{KL}(q_\phi(z|x)||p_\theta(z))}
\end{equation} 

The first term increases the likelihood of the training data being generated. The second term brings the approximate posterior close to the true prior, which has a closed form solution in the case of a vanilla VAE wherein both these distributions are gaussian. 

Therefore, the key highlight of VAE is that the inference network (the encoder) allows the inference of $\mathbf{q_\phi(z|x)}$ which can be used for representation learning.

\begin{figure}
    \centering
    \includegraphics[width=\columnwidth,scale=0.25]{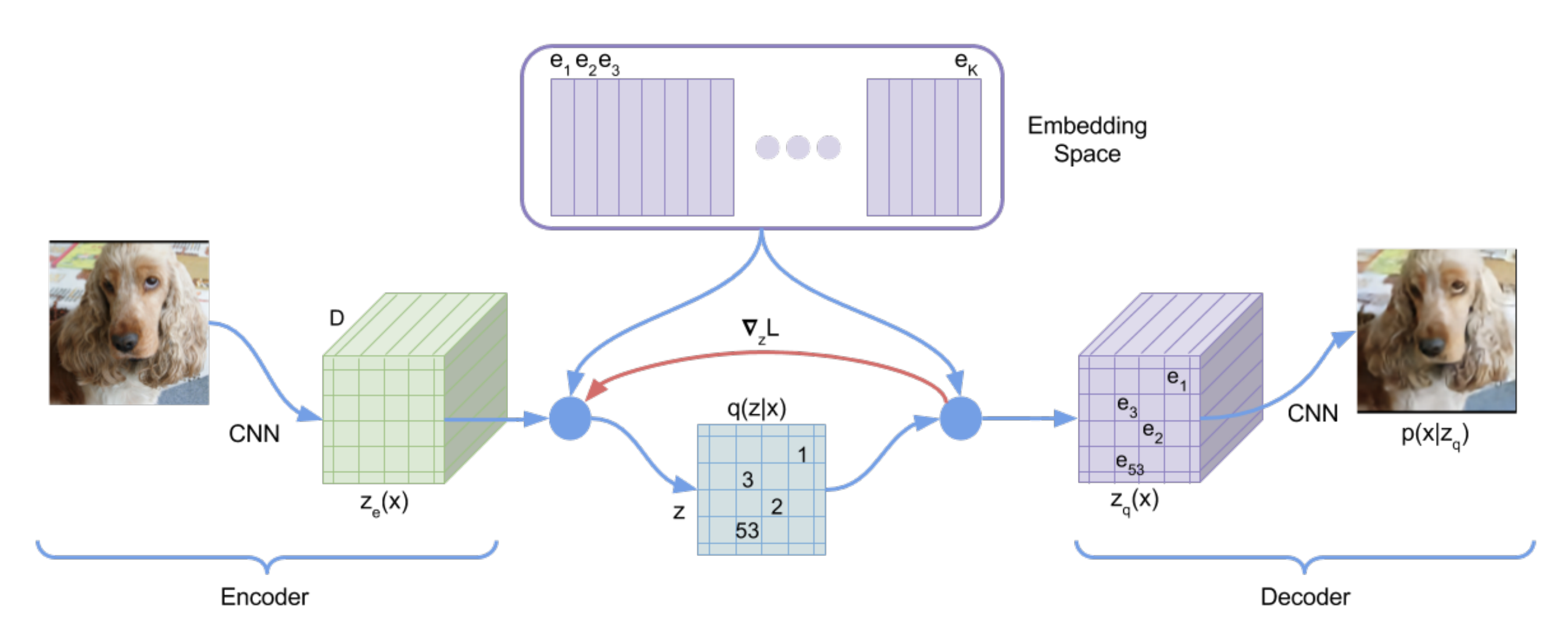}
    \caption{The VQ-VAE model. \cite{van2017neural}}
    \label{vqvae}
\end{figure}

\subsubsection{Vector Quantized Variational Autoencoder \cite{van2017neural}}
\cite{2014arXiv1411.4555V} shows that images are modelled better using discrete symbols. Most of the VAE models having a powerful decoder ignore the latent vectors. This situation is known as the posterior collapse. Keeping this in mind, \cite{van2017neural} introduced VQ-VAE, wherein the latents are discrete instead of continuous. Further, they show that the posterior collapse issue is solved. 

VQ-VAE adds a latent embedding space (also known as the dictionary) to the general VAE framework. Each embedding ${e \in R^{K \times D}}$, where K is the number of embeddings in the latent embedding space, and D is the dimension of each embedding. Here, the encoder is not used to model a gaussian distribution. Instead, its output, ${z_e(x)}$, is used to perform a nearest neighbour lookup on the vectors in the embedding space to obtain $\mathbf{z_q(x)}$. Hence, the continuous output vectors from the encoder are being quantized. This is known as vector quantization. Next, the decoder uses $\mathbf{z_q(x)}$ for the reconstruction task. Unlike a vanilla VAE, the posterior is a one-hot distribution over the embedding space.
\begin{equation}
q(z=k|x) = 
\begin{cases}
 1 &\text{for k = $argmin_j \norm{z_e (x) - e_j}_2$} , \\
 
 0 &\text{otherwise}
 \end{cases}
\end{equation}
Along with this, the loss function is the sum of ELBO loss, a dictionary learning term, and a commitment loss. The dictionary learning term moves the latent embeddings closer to the output of the encoder, and the commitment loss makes sure the encoder commits to the embedding by moving the output of the encoder to be closer to the chosen embedding from the dictionary. In the following equation, the stop gradient operation is defined as sg.
\begin{equation}
    L = \text{log } p(x|z_q(x)) + ||\text{sg}[z_e(x)]-e||_{2} ^2\  +\  \beta||z_e(x)-\text{sg}[e]||_{2} ^2
\end{equation} 
During training, the prior is chosen to be a uniform distribution over the embedding space. As a result, $D_{KL}(q(z|x)||p(z))$ becomes a constant equal to $\log(K)$. This indeed is independent of the network weights and hence, it is not included in the objective function, K turns into a hyper-parameter instead. Note that the vector quantization layer does not allow the use of backpropagation due to the arg-min operation. Instead, the gradient is directly copied from the decoder to the encoder, like the straight-through estimator. Fig. \ref{vqvae} gives a visual description of the VQ-VAE model.


\subsubsection{Generative Grasp Convolutional Neural Network (GGCNN) \cite{morrison2018closing}}

\begin{figure}
    \centering
    \includegraphics[width=\columnwidth,scale=0.7]{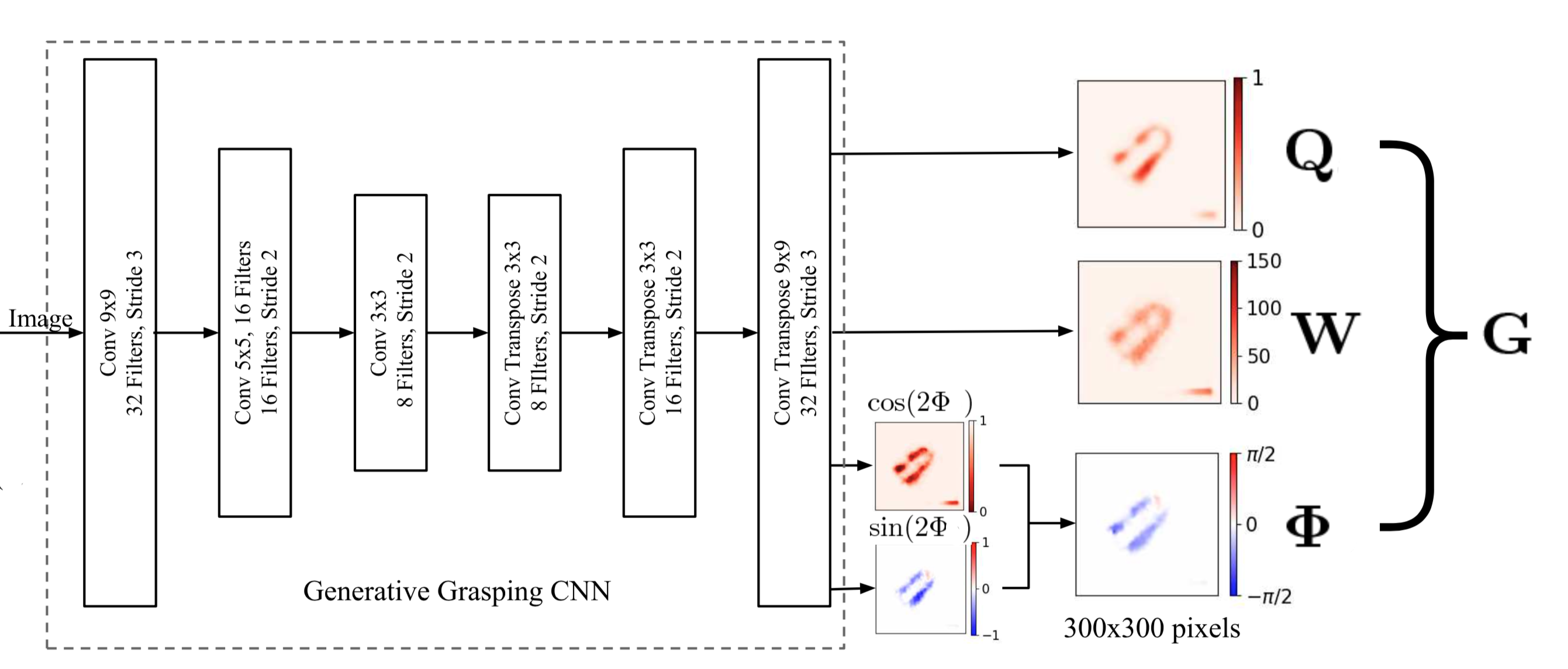}
    \caption{GGCNN network architecture. \cite{morrison2018closing}}
    \label{ggcnn}
\end{figure}

The Generative Grasp - CNN or the GGCNN model eliminates the need of a two staged pipeline to predict the grasp, thus, causing a huge reduction in both the number of parameters in the model, and execution time. This makes closed loop grasping feasible, causing the robot to visually detect changes in the environment as it reaches to grasp the object, and change the trajectory of the gripper accordingly.

Instead of sampling grasp candidates first and then following the pipeline, the GGCNN network directly predicts grasps on each and every pixel of the image. Given a 2D image \textbf{I}, the grasp $\mathbf{\tilde{g}}$ in the image can be represented as $\mathbf{\tilde{g}} = (\tilde{p},\tilde{w}, \tilde{\phi}, q) $, where $\tilde{p} $ denotes a pixel position $(i,j)$ which is the centre of the grasp rectangle in the image, $\tilde{w}$ and $\tilde{\phi}$ denote the gripper opening width and the angle of rotation of the gripper respectively. Let the image \textbf{I} be of dimension $H \times W$. The network architecture approximates a function \textbf{M} such that :
\begin{equation}
    \mathbf{G}^{3 \times H \times W} = M(\mathbf{I}^{H \times W})
\end{equation}
Here, \textbf{G} is the \textit{grasp map}, i.e. the set of all grasps over the image space and is denoted by $\mathbf{G} = (\mathbf{W}, \mathbf{\Phi}, \mathbf{Q})^{3 \times H \times W} $, where each of \textbf{W}, $\mathbf{\Phi}$ and \textbf{Q} are of dimension $ H\times W$, containing the values of $\tilde{w}$, $\tilde{\phi}$ and $\tilde{q}$, for each pixel $\tilde{p} = (i,j)$ of the image, where $0\leq i < H$ and $0\leq j < W$. 

The network architecture of the GGCNN model consists of multiple convolution layers stack against each other with varying kernels and strides as shown in Fig. \ref{ggcnn}.
\subsection{Proposed Approach}
\begin{figure*}
    \centering
    \includegraphics[scale=0.25]{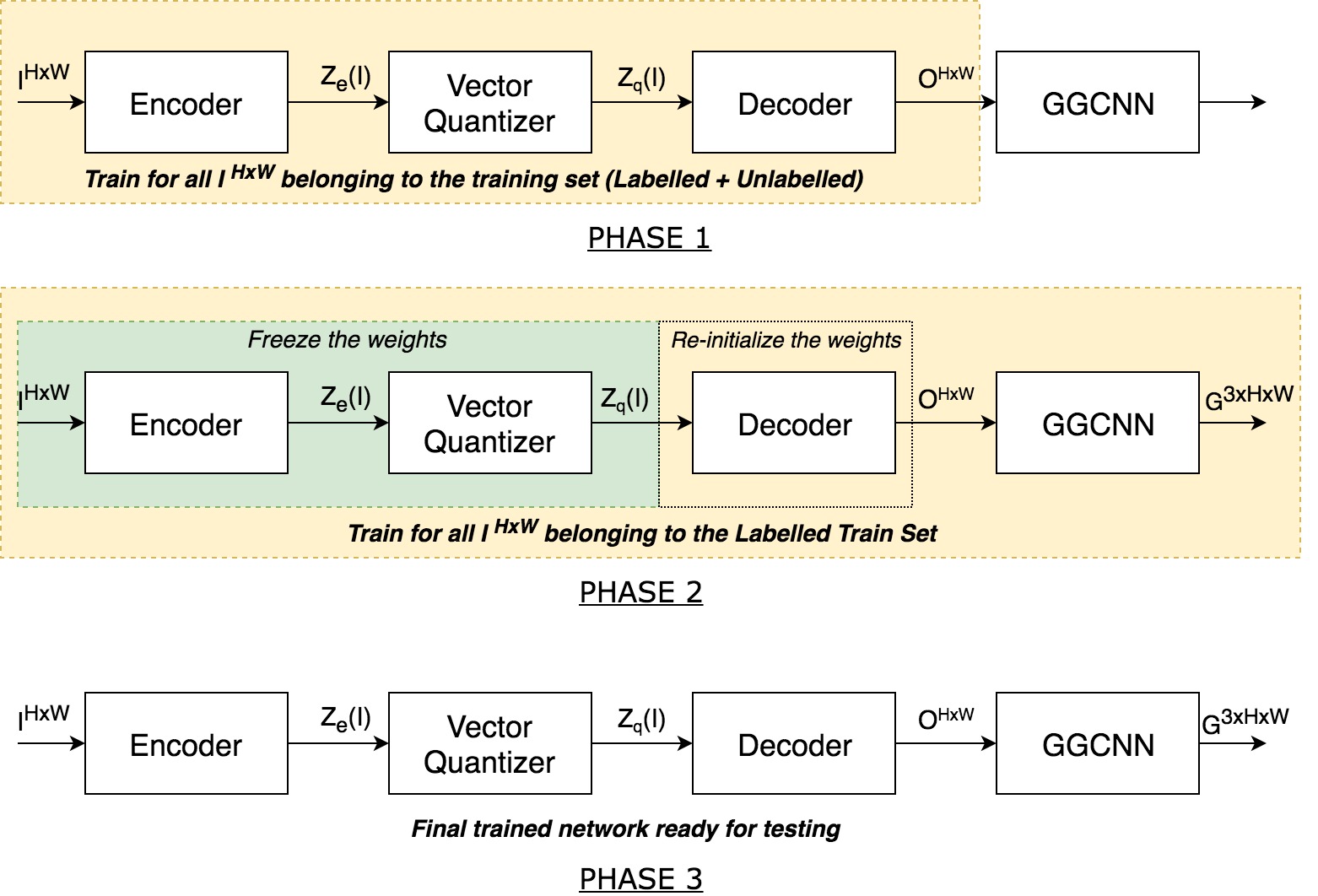}
    \caption{Proposed Approach.}
    \label{proposed}
\end{figure*}

Given a dataset with $n_1$ labelled training pairs of the form $(x, y)$, where $x$ is an RGB image and $y$ is the grasp vector, and $n_2$ unlabelled training vectors of the form $(x)$ with unknown $y$, we train a VAE on $n_1 + n_2$ images in the dataset. The trained VAE should capture important details which should help us in the supervised learning task that follows. We tried a vanilla VAE with normalizing flows, and a VQ-VAE to model the distribution of our training data. By observing the reconstruction quality, we came to the conclusion that VQ-VAE worked very well for our task. This decision was also based on the fact that VQ-VAE does not suffer from posterior collapse. The encoder and the quantization layers are then used to obtain $n_1$ latent vectors. These latent vectors are then clubbed with their corresponding $y$ values to obtain training pairs of the form $(z, y)$ for the supervised learning task, where $z$ is obtained by passing the corresponding $x$ through the encoder and the quantizer. These $(z, y)$ pairs are then used to train our modified GGCNN network. We altered the original GGCNN network by using the decoder architecture used in the VQ-VAE as the initial structure of the modified GGCNN. Intuitively, since the decoder architecture worked well in efficiently using the latent embeddings, it should benefit this task as well. Fig. \ref{proposed} gives a visual description of the proposed approach.

\section{Results and Analysis}
\begin{figure*}
\minipage{0.32\textwidth}
  \includegraphics[width=\linewidth]{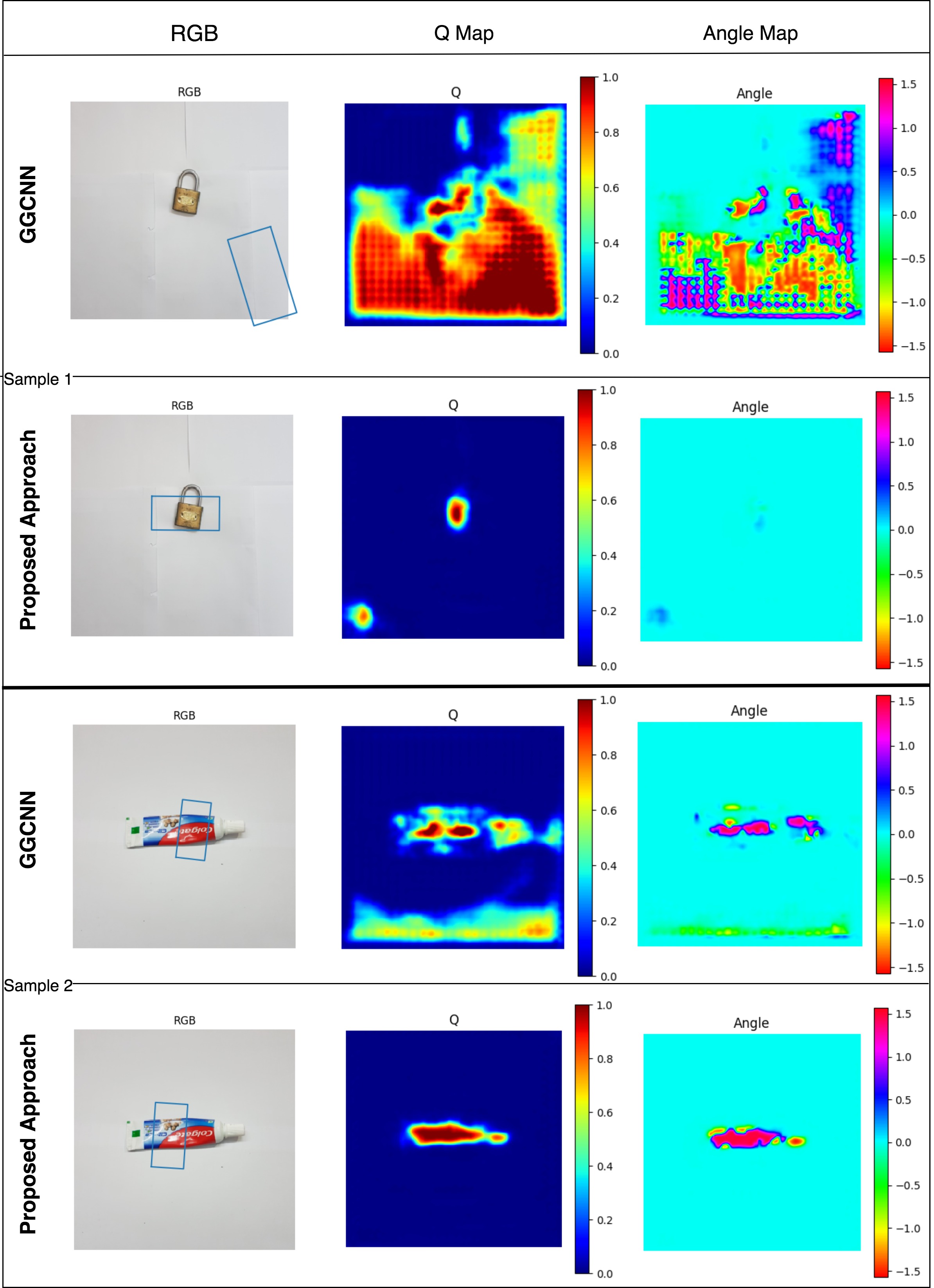}
\endminipage\hfill
\minipage{0.32\textwidth}
  \includegraphics[width=\linewidth]{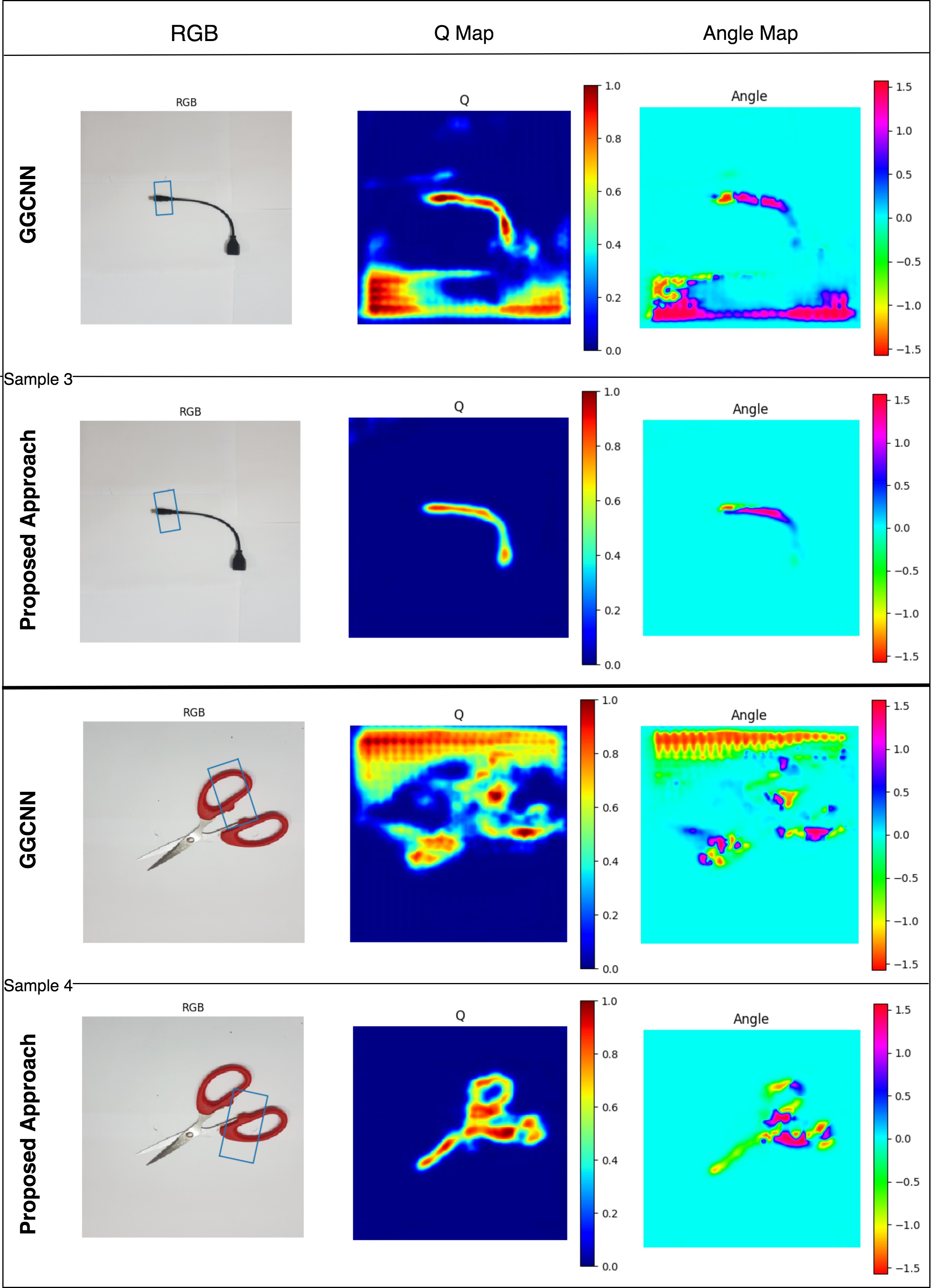}
\endminipage\hfill
\minipage{0.32\textwidth}
  \includegraphics[width=\linewidth]{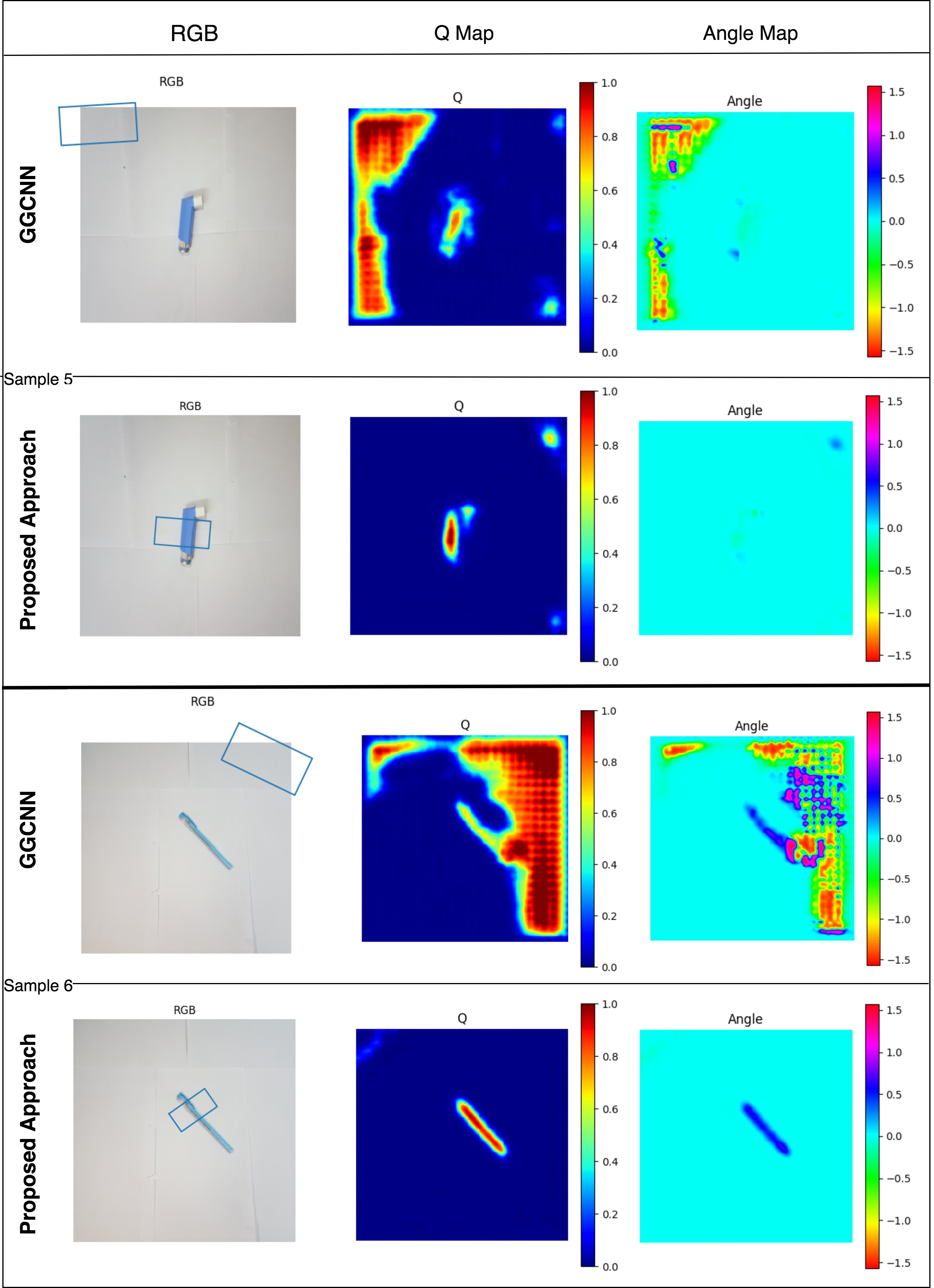}
\endminipage
\label{fig:restable}
\caption{A comparison between the grasp representations obtained from a GGCNN network and our proposed approach on different unseen objects (a lock, a tooth-paste, a cable, a scissor, an inhaler and a pen)}
\end{figure*}

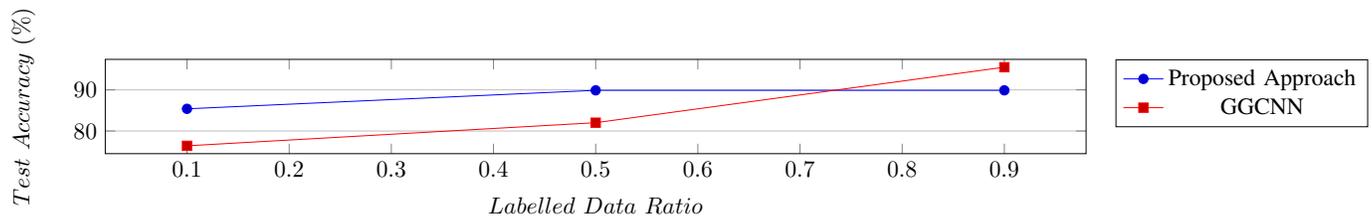
\begin{figure*}
\centering
\begin{adjustbox}{width=\textwidth}
\begin{tikzpicture}
    \begin{axis}[
    width=0.9\textwidth,
    height=3cm,
    ymajorgrids,
    ylabel=$Test$ $Accuracy$ $(\%)$,
    xlabel=$Labelled$ $Data$ $Ratio$,
    legend pos=outer north east
    ]
        \addplot table[x=in,y=vqvaeggcnn] {plotData.csv};\addlegendentry{Proposed Approach}
        \addplot table[x=in,y=ggcnn] {plotData.csv};\addlegendentry{GGCNN}
    \end{axis}
\end{tikzpicture}
\end{adjustbox}
\caption{A comparison between the GGCNN network and the proposed approach for varying amount of labelled data.} \label{fig:M1}
\end{figure*}

We use the Cornell Grasping Dataset \cite{60} to train our networks. The dataset contains 885 RGB-D images. Though the dataset is small, D Morrison et al. base the decision to use this dataset for the GGCNN network on the fact that it contains 5110 positive grasp labels, which benefits the prediction of the grasp map. Nevertheless, the dataset is augmented by performing transformations like rotation. As a preprocessing step, we divide every grasp rectangle in the Cornell Grasping Dataset into three parts and the centre rectangle is considered to be the position of the gripper’s centre. Although the Cornell Grasping Dataset contains negative labels as well, we follow the steps of the original GGCNN training procedure and consider any area other than the positive grasp rectangle to be an invalid grasp. 90\% of the dataset (10\% labelled data and 80\% unlabelled data) forms the train set, while the remaining 10\% forms the test set. We keep the labelled data percentage low to simulate the unavailability of sufficient labelled data for training.
For evaluating the predicted grasps, we use the Jaccard index as the evaluation metric. The predicted grasp is considered to be correct if the Jaccard index of the predicted grasp rectangle and the human-labelled grasp rectangle is greater than 25\%.  

The first experiment involves training the GGCNN network on only 10\% labelled images from the train set and evaluating on the test set. The second experiment is based on our proposed approach. Firstly, all the images in the train set are used as unlabelled images to train the VQ-VAE. Next, the weights of the encoder network and the vector quantization layers are frozen. Thereafter, the decoder network and the cascaded GGCNN network are trained on the labelled images in the train set. In the first experiment, we observe a test accuracy of $\mathbf{76.404\%}$. In the second experiment, however, the test accuracy is $\mathbf{85.3933\%}$, which shows the effectiveness of the proposed method.

Fig. 4 shows the Q Map and the Angle Map for multiple object samples, which are not present in the CGD. Color gradients have been used in place of numerical ranges. It is observed that both the networks produce good grasps when tested on images in the test set. However, the results are drastically different when tested on images outside the CGD. Though at times the GGCNN network is able to determine a good quality grasp rectangle (as shown in Fig. 4), the Q Map is far worse in comparison to the one produced by our model in all the cases. This proves that our model is able to generalize well.  Fig. \ref{fig:M1} plots the accuracy of the two approaches with varying amount of labelled data. Clearly, when the amount of labelled training data is low, our model performs much better than a neural network which utilizes only the labelled data in the training process for grasp detection.

\section{Conclusion and Future work}
In this work, we have presented a new approach for vision based robotic grasp detection. We have shown that semi-supervised learning approaches can be utilized to make use of limited labelled data available in the grasp detection domain. Further, we have shown that using a Vector Quantized Variational Autoencoder can help in extracting useful features for determining the grasp vector. Our experiments demonstrate that our new approach outperforms current approaches by a significant margin. Also, we have shown that our model was able to generalize as is evident by the results wherein it performed very well on images not present in the Cornell dataset, despite being trained on a small amount of labelled data. The results obtained from the GGCNN network, however, were not good in comparison.

In our approach, the labels had no influence on the estimation of the posterior distribution. Kingma et al. \cite{2014arXiv1406.5298K} focus on the classification problem and construct the posterior as $\mathbf{q(z,y|x)}$ instead and assume it to have a fully factorized form. Here, $\mathbf{q(z|x)}$ is modelled by the classifier. In future, our approach can be improved by constructing a better posterior which incorporates the GGCNN network and benefits from the labels as well. As a side-benefit, the whole training process would become end-to-end too.
\bibliographystyle{IEEEtran}
\bibliography{biblio}

\end{document}